\documentclass{article}

\usepackage[preprint]{neurips_2024}

\usepackage{caption}
\usepackage[utf8]{inputenc} 
\usepackage[T1]{fontenc}    
\usepackage{hyperref}       
\usepackage{url}            
\usepackage{booktabs}       
\usepackage{textcomp}
\usepackage{amsfonts}       
\usepackage{nicefrac}       
\usepackage{microtype}      
\usepackage{epsfig}
\usepackage{graphicx}
\usepackage{array}
\usepackage{amsmath}
\usepackage{array, makecell}
\usepackage{booktabs}       
\usepackage{amsfonts}       
\usepackage{amsthm}  
\usepackage{algorithm}
\usepackage{algpseudocode}
\usepackage[dvipsnames]{xcolor} 

\usepackage{newunicodechar}
\newunicodechar{′}{\ensuremath{\prime}}

\def\bmvaHangBox#1{
\begin{minipage}[t]{\textwidth}
\begin{tabbing} 
~\\[-\baselineskip] 
#1 
\end{tabbing}
\end{minipage}} 

\title{MOOSE: Pay Attention to Temporal Dynamics for Video Understanding via Optical Flows}

\newtheorem{assumption}{Assumption}
\newtheorem{definition}{Definition}
\author{%
  Hong Nguyen, Dung Tran, Hieu Hoang, Phong Nguyen, Shrikanth Narayanan \\
  Department of Computer Science\\
  Cranberry-Lemon University\\
  Pittsburgh, PA 15213 \\
  \texttt{hippo@cs.cranberry-lemon.edu} \\
}

\author{%
  Hong Nguyen$^{1}$ \quad
  Dung Tran$^{2}$  \quad
  Hieu Hoang$^{2}$ \quad
  \\
 \textbf{Phong Nguyen}$^{2}$ \quad
   \textbf{Shrikanth Narayanan}$^{1}$
   \vspace{1mm}
  \\
$^{1}$ University of Southern California \quad $^{2}$Hanoi University of Science and Technology \\
  \texttt{hongn@usc.edu}
}

\begin{document}

\maketitle

\begin{abstract}

Many motion-centric video analysis tasks, such as atomic actions, detecting atypical motor behavior in individuals with autism, or analyzing articulatory motion in real-time MRI of human speech, require efficient and interpretable temporal modeling.
Capturing temporal dynamics is a central challenge in video analysis, often requiring significant computational resources and fine-grained annotations that are not widely available. 
This paper presents MOOSE (Motion Flow Over Spatial Space), a novel temporally-centric video encoder explicitly integrating optical flow with spatial embeddings to model temporal information efficiently, inspired by human perception of motion. Unlike prior models, MOOSE takes advantage of rich, widely available pre-trained visual and optical flow encoders instead of training video models from scratch. This significantly reduces computational complexity while enhancing temporal interpretability. Our primary contributions includes ($1$) proposing a computationally efficient temporally-centric architecture for video understanding ($2$) demonstrating enhanced interpretability in modeling temporal dynamics; and ($3$) achieving state-of-the-art performance on diverse benchmarks, including clinical, medical, and standard action recognition datasets, confirming the broad applicability and effectiveness of our approach. Our code is publicly available at: \url{https://anonymous.4open.science/r/MOOSE-C055}
\end{abstract}

\section{Introduction}
Understanding temporal dynamics – the motion and relationships across video frames – is a core challenge in video analysis. 
Unlike static images, videos contain a time dimension along which objects move, interact, and change over time. Capturing these dynamics is often costly; well-annotated fine-grained datasets and sophisticated computing resources are typically accessible only to large corporations. For instance, encoding spatial features alone (without even considering temporal information) for one second of video ($30$ FPS) can require up to $30$ times more storage and computation than a single image.
The second big challenge comes from video data itself, which lacks fine-grained atomic annotations, a problem illustrated in VideoPrism \cite{videoprism24}. Existing data \cite{K400} focuses on describing the static content of the scenes rather than dynamic elements (e.g., actions), thereby downgrading a video model to an image model. For instance, the activity "playing golf" can be easily classified using an image-only encoder, while hand gestures such as "push" or "pull" may involve more temporal understanding).
\begin{figure}
\begin{tabular}{cc}
\bmvaHangBox{\includegraphics[width=0.25\linewidth]{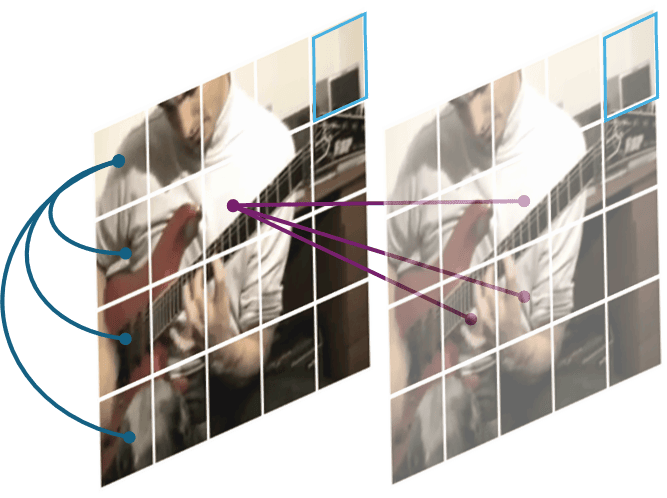}\\[-0.1pt]}&
\bmvaHangBox{\includegraphics[width=0.7\linewidth]{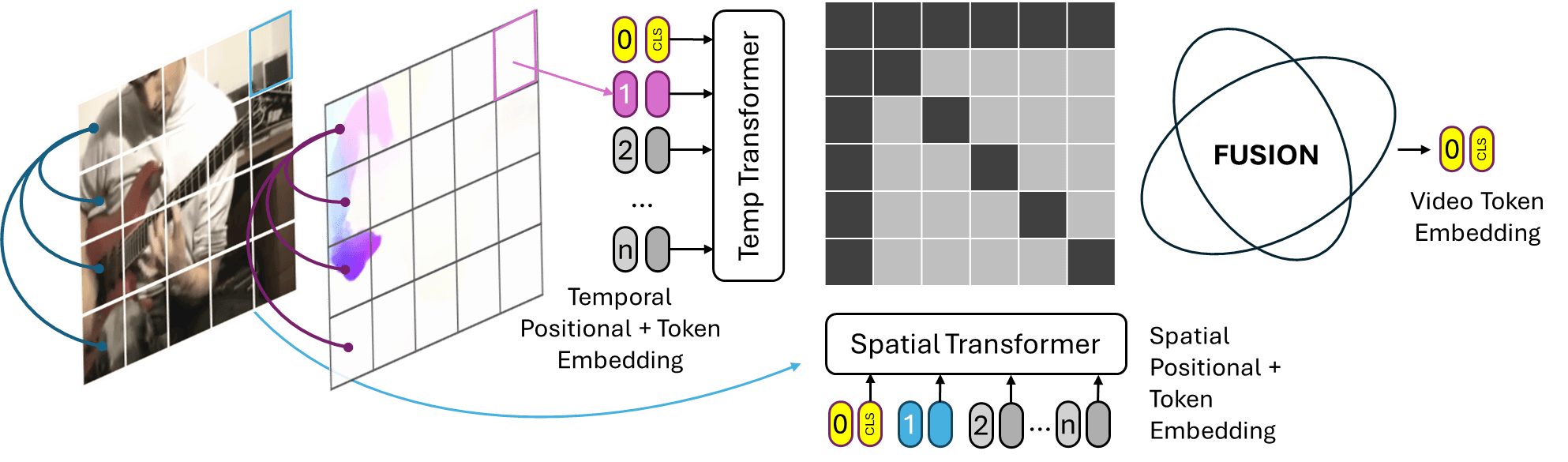}}\\
(a)&(b)
\end{tabular}
\caption{(a) Conventional attention-based methods perceive temporal dynamics with cross-spatial attention and (b) Our approach considers temporal dynamics as a series of vector fields (via optical flow). Flow vectors is grouped in patches to match image patches}
\label{fig:teaser}
\end{figure}


As humans, we perceive motion when two or more static images are presented in rapid succession \cite{cardullo2011human}. Recent deep learning architectures \cite{timesformer, vivit} have mimicked this phenomenon to encode temporal dynamics into embedding vectors by keeping track of the same object over frames. Pioneering work by Exner \cite{sigmud} demonstrated that the perception of motion can arise even when the spatial gap between the images is so small that they cannot be consciously distinguished as separate entities. 
This suggests that the brain does not simply infer motion by detecting positional changes over time, but rather perceives motion as a direct and intrinsic sensory experience, much like the perception of color or sound.
Inspired by Exner's suggestion, we develop a video model that embeds temporal dynamics as a distinct modality (with visual modality) via optical flow. Optical flow captures the patterns of apparent motion of objects, surfaces, and edges in a visual scene caused by the relative motion. 

Recent advances \cite{flowsvm07, park2024conv3d, papadakis2024multi} demonstrate that integrating optical flow into temporally-aware architectures leads to improvements in both accuracy and interpretability across a range of motion-centric tasks. This approach has proven particularly valuable in domains where subtle motion cues carry significant semantic weight. For instance, in medical and clinical applications, motion-sensitive models have been used to track articulator movements in real-time MRI \cite{75speaker}, quantify repetitive behaviors in autism assessments, and analyze joint attention in caregiver–child interactions \cite{avasd, childplay, MMASD}. These scenarios often involve low-resolution or noisy data, where temporal cues can be more informative than static spatial features. Thus, motion modeling—augmented by optical flow—provides a generalizable and effective foundation for a broad class of video understanding problems, spanning both general-purpose and domain-specific applications.
This paper introduce a novel video understanding framework that prioritizes temporal modeling while maintaining computational efficiency. By focusing on temporal features rather than spatial complexity, our architecture significantly reduces resource demands without compromising performance.
Another central contribution is the improved interpretability of temporal dynamics within the model. Our design allows for more transparent analysis of how motion and visual influence predictions via cross-attention.
Finally, we validate the versatility and robustness of our approach through extensive evaluation on a wide range of datasets. 

\section{Related Works}
Various techniques have been developed to capture temporal dynamics in video data. These models aim to encode the spatio-temporal context of a video into a latent embedding that can be used for tasks like action recognition. We can categorize these models as explicit or implicit temporal dynamics depending on whether or not they can explain or visualize motion cues across frames.

\noindent
\textbf{Implicit Temporal Dynamics} 
Several approaches \cite{Rev-ViT, mvit, x3d, conssl, c2d, i3d, batchtrans} treat videos as 3D entities spanning time, width, and height. For instance, VideoMAE \cite{videomae, videomaev2} reconstructs 3D video clips from inputs masked in the time domain while VideoSwin \cite{videoswin} partitions video clips into 3D patches and embeds them using standard transformer mechanisms. However, these approaches learn spatiotemporal representations jointly without regard to the unique aspects of temporal information.

\noindent
\textbf{Explicit Temporal Dynamics} 
The SlowFast \cite{slowfast} model’s representation of temporal dynamics is essentially a disentanglement of low and high frame rate. 
TimeSformer \cite{timesformer} and ViViT \cite{vivit} rely on cross-frame attention to explicitly identify temporal relations (with TimeSformer separating time and space in attention computation and ViViT providing options for factorization).

\noindent
\textbf{Optical Flows as Temporal pathway}
    Recently, several approaches \cite{SevillaLara2017OnTI, motionformer} have studied optical flow as a condition vector to embed temporal information implicitly via cross-frame attention. Specifically, each patch of the current frame establishes attention weights toward patches of adjacent frames (before and after). By doing this, the final video representation will contain time and space redundancies, and motion dynamics can easily be explained via attention weights.
    Another representation of the temporal dimension is optical flow \cite{dong2024memflowopticalflowestimation}, which computes the motion between consecutive frames, thereby characterizing temporal changes and movement. Some works \cite{batchtrans} adopt optical flow for action recognition tasks, yet the interplay between spatial and temporal aspects is not clearly represented but behave as a black-box. 
    
MOOSE is inspired by both attention-based and flow-based video encoders, combining their strengths of intuitive explainability via optical flows and rich pretrained representation from vision transformer. We show that the interplay between spatial patches and optical flow patches provides meaningful insights about video.

\begin{figure*}
\begin{center}
\includegraphics[width=\linewidth]{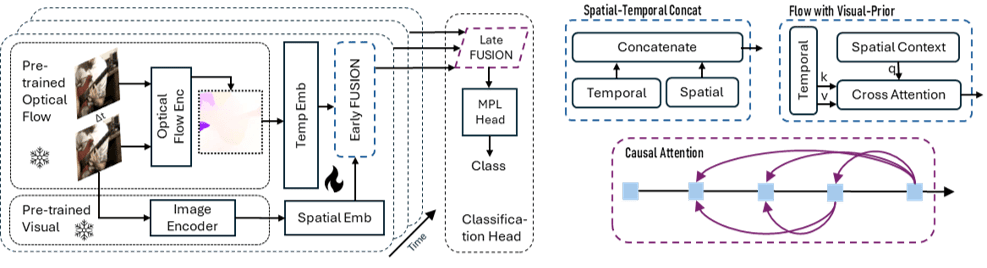}
\end{center}
   \caption{The proposed video model for atomic activity classification treating time and space as two distinct modalities (visual and motion, respectively). To save computational resources while preserving performance, we use visual and optical flow pre-trained models.}
\label{fig:system}
\end{figure*}

\section{MOOSE Video Encoder}
We consider a video of duration $n \times \Delta t$ as a time series comprising
$n$ video units, where each unit represents not just a conventional 2D frame, but rather a 3D construct, with the third dimension capturing flow velocity information. Each unit is processed through spatial and temporal pathways, which will be presented in sections \ref{temppath}. The spatial pathway captures contextual visual information, while the temporal pathway, represented by optical flow, encodes the movement of visual patterns. While optical flow focuses on motion, it can not serve as an independent modality because the motion vectors it calculates are inherently tied to the spatial context of pixels within an image frame. For instance, knowing that a group of pixels moving in a particular direction corresponds to a car requires a contextual understanding of the objects corresponding to the optical flow vectors. Therefore, we discuss early fusion of optical flow with spatial context within a video unit in section \ref{spatempfusion}. Since optical flow can only capture temporal dynamics over a short time duration, it may be insufficient for video-level tasks that span multiple seconds or even minutes. Section \ref{timeaggr} will discuss space-time aggregation over video-level classification.
\begin{definition}[Video]
A video is a series of space-time units: each unit contains a spatial (one frame) and temporal (velocity field of optical flow at that frame during $\Delta t$) pathway. If we have T time points or n frames, a video would
therefore be represented by an element $z \in \mathbb{R}^{T \times 3 \times W \times H}$ where $T = n \times \Delta t$ and the long-term temporal pathway is the trajectory of flows.
\end{definition}

\begin{assumption}[Optical flows existence]
\label{assumopticalflow}
 Changes in the image intensity of the input frames during $\Delta t$ are only due to the transition of
the local image intensity. Context information is assumed to be preserved within frames.
\end{assumption}

\noindent
\textbf{Input.} The MOOSE encoder takes as inputs video clips $z \in \mathbb{R}^{T \times C \times W \times H}$ sampled at rate $r = 1/\Delta t$ from the original video with sampling period $\Delta t$ and outputs a video embedding vector $e \in \mathbb{R}^{L}$ where $T, C,$ and $L$ are the number of frames, color channels, and the length of the embedding vector respectively. 
To ensure the validity of Assumption \ref{assumopticalflow}, the sampling period $\Delta t$ was selected to be sufficiently small.


\noindent
\textbf{Spatial Pathway.} 
Vision Transformers (ViTs) have emerged as a powerful architecture for processing visual data, drawing inspiration from the success of transformers in natural language processing. Utilizing ViTs, we encode each image in video clips as spatial embeddings. During time period $n\Delta t$, consecutive frames present spatial redundancies, as spatial context is similar among frames, due to assumption \ref{assumopticalflow}. As a result, spatial embeddings are extremely similar in terms of cosine distance. To differentiate these frames in an atomic sense, temporal pathways are introduced in Section~\ref{temppath} to encode video units.


\noindent
\textbf{Sampling rate.} The sampling rate refers to the frequency at which frames are selected from the original frame sequence for model input. An increase in the sampling rate generates differences in spatial context and introduces uncertainties in the temporal domain. 
In this study, we primarily consider frames sampled at a low rate, typically $1/8$
of the original frame rate, with an input size of $8 \times 224 \times 224$, unless otherwise specified.



\subsection{Temporal Pathway}\label{temppath}
In the temporal pathway, a sequence of optical flow features extracted from consecutive frames are fed into a transformer-based encoder. Here, the self-attention mechanism can be effectively applied to flow vectors, as shown in Fig \ref{fig:system}(b), allowing the model to understand motion patterns within each frame.


\noindent
\textbf{Optical flows estimation} We first estimate optical flows $v \in \mathbb{R}^{(T-1) \times C_t \times W \times H}$ for each pair of frames in clip $X \in \mathbb{R}^{T \times C \times W \times H}$, where $C_t$ represents the vertical and horizontal directions of changes, and $C_t$ equals 2 for 2D optical flow.

\begin{definition}[Velocity field of optical flow]
A velocity field is a vector field containing every optical flow vector $v \in \mathbb{R}^{2 \times W \times H}$ that describes the magnitude of change in horizontal and vertical directions.
\end{definition}

\noindent
\textbf{Temporal patches and self-attention}
Previous studies have modeled explicit temporal dynamics between frames through attention mechanisms by capturing changes in location of the same object. In contrast, this paper treats the temporal pathway as a dependent modality of the spatial domain, and utilizes optical flow to represent it. We treat optical flow as an image input for ViT where each patch embedding is flow patches and the \texttt{cls\_token} is a temporal classification token. The number of channel equals 2. The connection between spatial and temporal dimensions is represented through corresponding patches between 2D-image and 2D-flow vectors. We divide the flow matrix into smaller patches and embed each to a representation vector using a vision transformer. Each patch has the same size as the visual patches, so each spatial patch will have corresponding flows. 
Self-attention weights are
computed using a dot product on temporal patch embeddings and \texttt{cls\_token}. We apply a Vision Transformer to optical flow as done on images and split each optical flow matrix $v \in \mathbb{R}^{C \times W \times H}$ into N non-overlapping patches. Each patch of dimension $(C, P, P)$ is considered as a flow-patch embedding in which $P$ is the patch size. Temporal patches are flattened and summed with positional embeddings: $z_{p,t} = E x_{p,t} + pos$.


\noindent
\textbf{Arrow Masking} Transformers use masking in attention mechanisms to control what information a token can attend to during training and inference. Masking is crucial for both bidirectional and autoregressive models to ensure proper learning dynamics. In autoregressive models such as GPT, tokens should not "see" future words during training. As for our model, each visual patch has a corresponding motion patch embedding that encodes temporal information for the spatial content. Each visual patch should only attend to its likely movement. Moreover, the spatial and temporal \texttt{cls\_token} should have an overall view of other patch embedding spaces to make classification decisions. Figure \ref{fig:teaser}(b) shows our masking approach in which the diagonal represents pairwise attention between corresponding spatial-temporal patches and the first row and column represent attentions of \texttt{cls\_token} towards both modalities.

\subsection{Spatial-Temporal Fusion for Video Units}\label{spatempfusion}
The purpose of early fusion is to encode spatial-temporal vectors within video units. 
Optical flow is informative but not a standalone modality.
As shown in Figure~\cite{fig:system}, we proposed several architectures to fuse temporal and spatial features. Each of the presented approaches represents a different hypotheses regarding human perception of video.


\noindent
\textbf{Approach 1: Video Unit is Image Embedding with Flow-Prior} We represent video embeddings as traditional list of image embeddings but each will be injected with flow information. By doing this, we leverage current image encoders with additional information around tentative changes in movement
\begin{align}
    \text{e}_\textit{z,t}^\text{flow-prior} & = \text{e}_{s,t} + \text{Attention}(Q_s,K_f,V_f) = \text{e}_\text{s,t} + \text{Softmax}(\frac{Q_sK_f^T}{\sqrt{d_k}})V_f
\end{align}
in which $\text{e}_\text{z,t}$, $\text{e}_\text{s,t}$ is embedding vectors of video units and images (spatial) at time point $t$. 
In cross-attention, the $Q_s = X_s W_s$ is conditional part draw from spatial encoder where $X_s$ are the spatial embeddings (including both classification token and patches) and $W_s$ is a weight matrix.

\noindent
\textbf{Approach 2: Video Unit is Flows Embedding with Visual-Prior} Consider models that have input as optical flows, each flow-patches will not have meaning without spatial context. As such, we attach spatial context to each flow-patches via cross-attention.
\begin{align}
    \text{e}_\text{z,t}^\text{visual-prior} & = \text{e}_{f,t} + \text{Attention}(Q_f,K_s,V_s) = \text{e}_\text{f,t} + \text{Softmax}(\frac{Q_fK_s^T}{\sqrt{d_k}})V_s
\end{align}

\noindent
\textbf{Approach 3: Video Unit is Joint Visual-Motion Embedding} This model considers visual embeddings and optical flow embeddings as two independent modalities, although optical flow is generated experimentally from images. We concatenate the two modalities and jointly back-propagate bidirectional cross-attention \cite{kitaev2020reformer}
\begin{align}
    \text{e}_\text{z,t} & = \left[\text{e}^\text{flow-prior}_{z,t}, \text{e}^\text{visual-prior}_{z,t}\right]  = \text{Bidirectional-Attention}\left(\text{e}_{s,t}, \hspace{2pt} \text{e}_{f,t}\right)
\end{align}

\begin{table}
\small
  \caption{Top-1 and top-5 accuracies of various video models on the Kinetics-400 (K400) and Something-Something v2 (SSv2) datasets. '-' indicates pretrained model was not provided by the authors or otherwise unknown. GPU hours are normalized to the scale of Tesla V100 using approximate relative compute ratios: V100 $\simeq1\times$, A6000 $\approx 1.4\times$, A100 $\approx 2.5\times$}
  \label{sample-table}
  \centering
  \begin{tabular}
  {lcccccccc}
    \toprule
    &&&&&\multicolumn{2}{c}{K400}&\multicolumn{2}{c}{SSv2}\\
    \cmidrule(r){6-7}\cmidrule(r){8-9}
    Model     & \makecell{Pretrained \\ dataset} & \makecell{GPU \\ hours}  & Param   & TFLOPs   &Top 1 &Top 5 &Top 1 &Top 5\\
    \midrule
    \multicolumn{4}{l}{\textit{3D CNNs}} \\
    SlowFast & ImageNet-1K  & 3840  & 34.57M & 16.52G  &62.81  &82.67 &60.35 &88.00\\
    I3D     & ImageNet-1K & 1440 & 28.04M & 34.28G    &67.31 &86.19 &-&-  \\
    C2D     & ImageNet-1K & -& 24.33M & 37.46G    &65.65 &84.86 &-&- \\
    \midrule
    \multicolumn{4}{l}{\textit{Temporal as Cross-frame Attention Approaches}} \\
    TimeSformer     & ImageNet-1K & 416 & 121.57M & 196.05G    &68.72	&88.83	&63.88	&89.3    \\
    ViViT-B     & ImageNet-21K       & - & 0.73M & 270.44G &51.3	&74.21 &-&-  \\
    VideoMAE     & None      & 1218 & 0.72M & 135.17G  &59.18	 &80.79	& 49.85	& 79.35 \\
    \midrule
    \multicolumn{4}{l}{\textit{Temporal as Optical Flow Approach}} \\
    MOOSE (Ours)     & \makecell{ImageNet-1K \\ \& Sintel}  & 132 &  104.70M & 297.3G  &\textbf{70.84}	&\textbf{89.56} &\textbf{65.23}	&\textbf{89.96}    \\
    \bottomrule
  \end{tabular}
\end{table}

\vspace{-5pt}
\subsection{Aggregate spatial-temporal units} \label{timeaggr}
Because optical flow captures motion only over very short time intervals, it may not be sufficient for tasks requiring modeling longer durations, such as classifying entire videos (spanning seconds) or understanding narratives (unfolding over minutes). To address this, this section will explore methods for aggregating information across both space and time for video-level classification.
To capture the dynamic evolution within video segments represented by embeddings from time t to t+n, a simple approach like averaging these embeddings proves insufficient as it collapses the temporal dimension, thereby losing crucial information about motion and sequence. Instead of this naive aggregation which sacrifices temporal resolution, our proposed method employs causal self-attention across the sequence of embeddings. 
\vspace{-5pt}
\begin{align}
    \text{e}_\text{z}^\text{clip-level} & = \text{Causal-Attention}\left(\left[\text{e}_{z,t}, \text{e}_{z,t+1}, ..., \text{e}_{z,t+n}\right]\right)
\end{align}
By applying self-attention with a causal mask (ensuring that predictions for time t$\prime$ only depend on inputs from t up to t$\prime$), the model learns the intricate temporal dependencies between frames. This allows it to explicitly model and learn the trajectories of objects or points of interest as they move and change throughout the video clip, preserving the vital sequential information that averaging discards.

\begin{table}[t]
\small
\centering
    \begin{minipage}[t]{.50\linewidth}
        \caption{Video-level accuracy of K400 dataset on different fusion approaches} 
        \vspace{4pt}
        \begin{tabular}[t]{lccc}
        \toprule
        Fusion Mode & TFLOPs  &Top 1 &Top 5 \\
        \midrule
        TimeSformer & 196.05G
        &68.72	&88.83\\
        Bidirectional Concat & 297.3G & \textbf{70.84} & \textbf{89.56} \\
        Visual-Prior Cross-Att & 299.26G & 55.73 & 78.79 \\
        Motion-Prior Cross-Att & 293.26G & 64.37 & 85.16\\
        \bottomrule
        \label{tab:fusionmodes}
        \end{tabular}
    \end{minipage}%
    \hfill
    \vspace{-3pt}
        \begin{minipage}[t]{.48\linewidth}
      \centering
        \caption{Video-level accuracy of K400 on different temporal aggregation approaches}
        \vspace{4pt}
        \begin{tabular}[t]{lccc}
        \toprule
        \makecell{Video Units \\Aggregation} &\makecell{Param}   &Top 1 &Top 5 \\
        \midrule
        Mean &101.01M &64.19  &88.19\\
        Causal &104.70M&\textbf{70.84}&\textbf{89.56}\\
        Mamba \cite{mamba} &101.01M &66.30  &86.52\\
        \bottomrule
        \label{tab:tenoaggr}
        \end{tabular}
    \end{minipage}%
    \hfill
    \begin{minipage}[t]{\linewidth}
      \centering
        \caption{Accuracy on other datasets}
        \vspace{4pt}
        \begin{tabular}[t]{lcccccc}
        \toprule
        Model & Test Crops &Params &rtMRI-Pho &AV-ASD &HAA500 &UCF-101 \\
        \midrule
        TimeSFormer &1 $\times$ 1& 121.57M &\textbf{85.27} &29.41 &68.09 &\textbf{98.79}\\
        MOOSE (Ours) &1 $\times$ 1&101.01M &77.62 &\textbf{31.99} &\textbf{72.42} &98.52\\
        \bottomrule
        \label{tab:otherdatasets}
        \end{tabular}
    \end{minipage}%
\end{table}

\section{Experiments}
\subsection{General Settings}
We use ViT architecture for both visual and motion embedding but with different model sizes. The optical flow estimator RAFT \cite{raft} was chosen, that is pre-trained on the Sintel \cite{sintel} dataset. All pre-trained weights of spatial ViT and RAFT are frozen. Trainable weights include Temporal ViT, classification head, and fusion module. We trained our model on a single GPU A6000 with 48GB of RAM. Following the setting of \cite{timesformer}, we use an SGD optimizer with weight decay of 1e-4. Spatial ViT is pretrained on Imagenet-1K using self-supervised methods DIONv2 as in \cite{darcet2023vitneedreg, dinov2}. The patch size is 14 $\times$ 14 for DINOv2. Unless differently indicated, we use clips of size 8 × 224 × 224, with frames sampled at 8 frames/s.

\textbf{Datasets}
We evaluate our method on two widely used and complementary video action recognition benchmarks: Kinetics-400 (K400) \cite{K400} and Something-Something V2 (SSv2) \cite{ssv2}. K400 is a large-scale dataset consisting of approximately 240,000 training videos spanning 400 human action classes, such as “playing guitar” or “riding a bike.” The actions are typically coarse-grained and heavily reliant on scene and object context, making it a standard benchmark for evaluating spatial and visual-based modeling capabilities. In contrast, SSv2 contains around 220,000 short video clips across 174 fine-grained action classes, such as “pushing something from left to right” or “pretending to put something on a surface.” These actions require understanding object manipulation, motion patterns, and temporal ordering, making SSv2 particularly well-suited for evaluating temporal reasoning and motion-sensitive video encoders. Together, these datasets provide a comprehensive testbed to assess both spatial and temporal aspects of videos.
To assess generalizability, we test our method on diverse datasets, beyond standard benchmarks. such as the 75-speaker rt-MRI data of human vocal articulation \cite{75speaker}, HAA500 \cite{haa500}, UCF-101 \cite{ucf101}, AV-ASD \cite{avasd}.








\subsection{Performance comparisons to State-of-the-Art}
\textbf{Accuracy.}
Table 1 presents a comprehensive comparison of our proposed method, MOOSE, with several representative video action recognition models across two benchmark datasets: Kinetics-400 (K400) and Something-Something V2 (SSv2). We categorize the baselines into three groups: (1) 3D CNNs, including SlowFast, I3D, and C2D, which rely on spatiotemporal convolutions; (2) Transformer-based approaches, such as TimeSFormer and ViViT, which utilize cross-frame attention mechanisms; and (3) self-supervised methods, like VideoMAE, which learn temporal structure via reconstruction objectives. All models are evaluated under a unified input setting of 8 frames with an 8×8 frame $\times$ sampling rate, and GPU hours scaled to Tesla V100 equivalence. For consistency, we use publicly reported numbers and scale GPU hours for A6000 and A100 based on relative throughput.
Our method, MOOSE, leverages visual encoder on pretrained on ImageNet-1K and optical flows extracted via models  pretrained on Sintel. By explicitly modeling optical flow and temporal dynamics in a lightweight architecture, MOOSE achieves state-of-the-art performance on Kinetics-400 (Top-1: 70.84\% and Top-5: 89.56\%) and SSv2 (Top-1: 65.23\%, Top-5: 89.96\%), outperforming heavy models such as VideoMAE and TimeSFormer.

Importantly, MOOSE achieves these results with only 132 GPU hours, which is ~9× lower than SlowFast and significantly more efficient than most transformer-based architectures. This efficiency highlights MOOSE’s suitability for scenarios where computational resources are limited or where scaling to larger datasets is required. The results confirm the effectiveness of explicitly incorporating motion representations through optical flow, particularly in datasets like SSv2 that demand fine-grained temporal reasoning.

\noindent
\textbf{Spatial and Temporal Resolution Scaling.}
We evaluate the GPU memory usage of MOOSE and TimeSFormer during training under varying spatial and temporal resolutions. Figure \ref{fig:tsneplot}(b) shows memory consumption with increasing input image sizes at a fixed batch size of 1. MOOSE consistently requires less memory than TimeSFormer, using about 30\% less at 448×448 resolution and remaining within the 48GB GPU limit even at 1024×1024, where TimeSFormer fails to run. Figure \ref{fig:tsneplot}(c) presents results for increasing temporal resolution (i.e., number of input frames) at a batch size of 16. MOOSE again scales more efficiently, supporting up to 128 frames without exceeding memory constraints, while TimeSFormer reaches the 48GB ceiling at only 16 frames. These results demonstrate that MOOSE is substantially more memory-efficient across both spatial and temporal dimensions, making it better suited for high-resolution inputs and long video sequences in memory-constrained training environments.
\subsection{Explainability compared with State-of-the-Art}
In this sub-section, we present visual explainability of our model compared with explicit SOTA video encoders.



\begin{figure*}[t]
\begin{center}
\includegraphics[width=\linewidth]{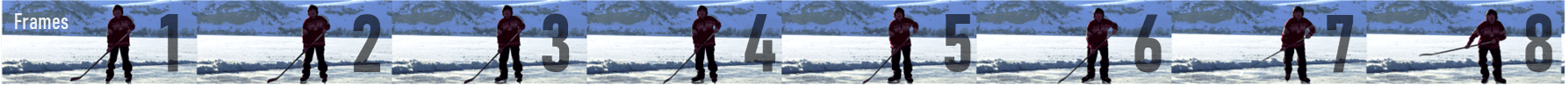} \\
\includegraphics[width=\linewidth]{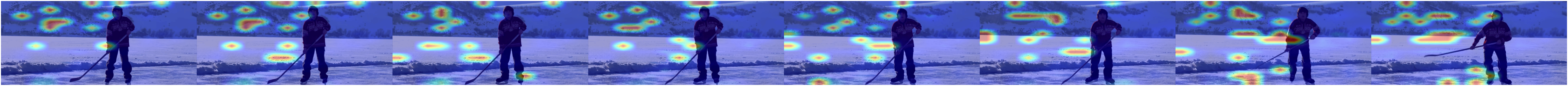} \\
\includegraphics[width=\linewidth]{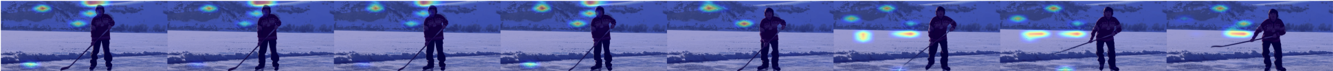} \\
\includegraphics[width=\linewidth]{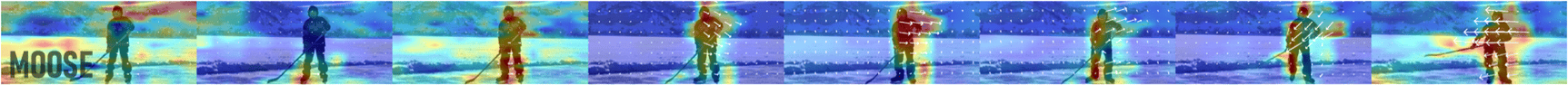}
\end{center}
   \caption{Eight continuous frames of original image followed by attention heatmap of (from top down) timesformer, vivit and moose}
\label{fig:vidmodelsCAM}
\end{figure*}

\begin{figure*}[t]
\begin{center}
\includegraphics[width=\linewidth]{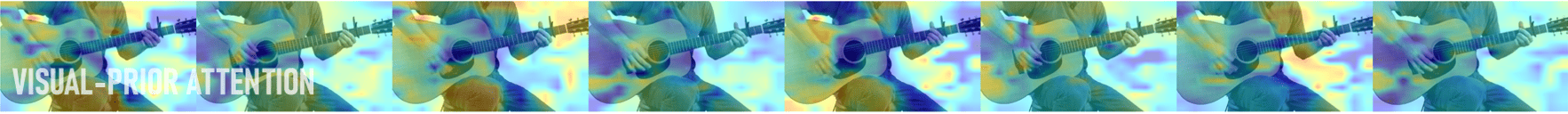} \\
\includegraphics[width=\linewidth]{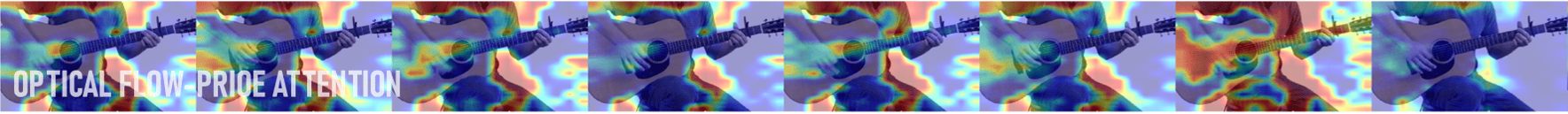} \\
\includegraphics[width=\linewidth]{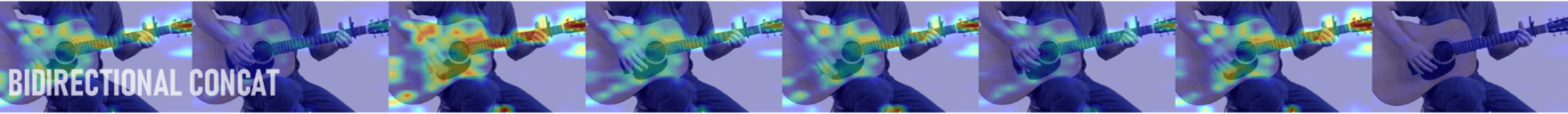} 
\end{center}
   \caption{Qualitative comparison of attention heatmaps over eight consecutive video frames for three model configurations: Visual-Prior Attention, Optical Flow-Prior Attention, and Bidirectional Concatenation}
\label{fig:vidmodesCAM}
\end{figure*}
\begin{figure*}[t]
\begin{center}
\includegraphics[width=0.49\linewidth]{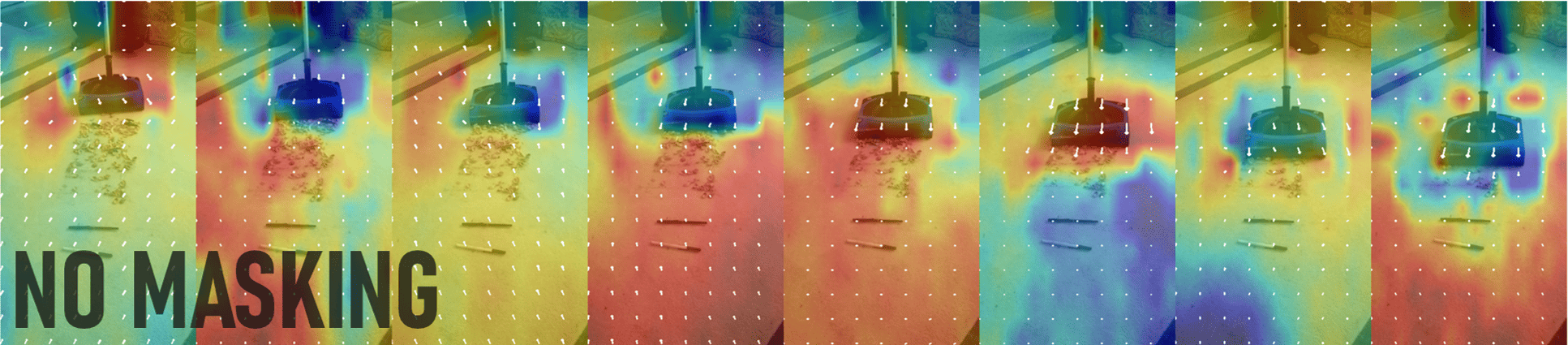} 
\includegraphics[width=0.49\linewidth]{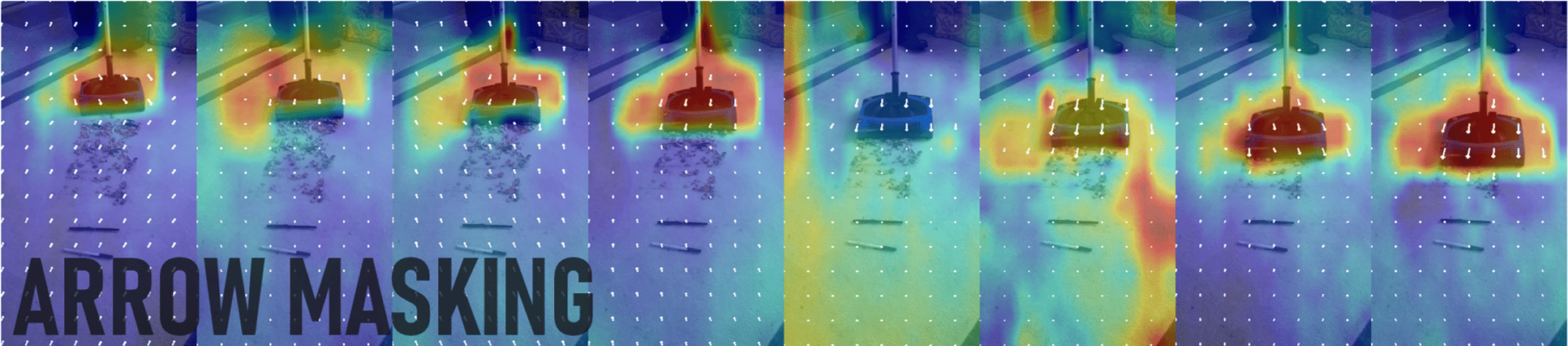}
\end{center}
   \caption{Attention heatmap of flow space between no masking (left) and arrow masking (right)}
\label{fig:short}
\end{figure*}

\noindent
\textbf{Spatiotemporal Attention Visualization}
To qualitatively evaluate how different models attend to relevant regions over time, we visualize the spatiotemporal attention maps across eight consecutive video frames in Figure \ref{fig:vidmodelsCAM}. The first row shows the original input frames, followed by attention heatmaps generated by TimeSFormer, ViViT, and MOOSE. As shown, both TimeSFormer and ViViT exhibit dispersed and sometimes inconsistent attention, often failing to focus precisely on the human actor or the dynamic motion regions relevant to the action.
In contrast, MOOSE’s optical flow attention consistently highlights the motion trajectory and silhouette of the actor across frames, indicating a strong temporal grounding. Moreover, the MOOSE temporal attention module further enhances focus on the actor’s limbs and moving objects (e.g., skis, poles). This refined localization illustrates MOOSE's superior ability to capture dynamic motion and discriminate between relevant foreground actions and static background, which is critical for complex action recognition tasks.

\begin{figure}[t]
\captionsetup{font=small}
    \centering
    \begin{minipage}{0.56\textwidth}
    \includegraphics[width=\linewidth]{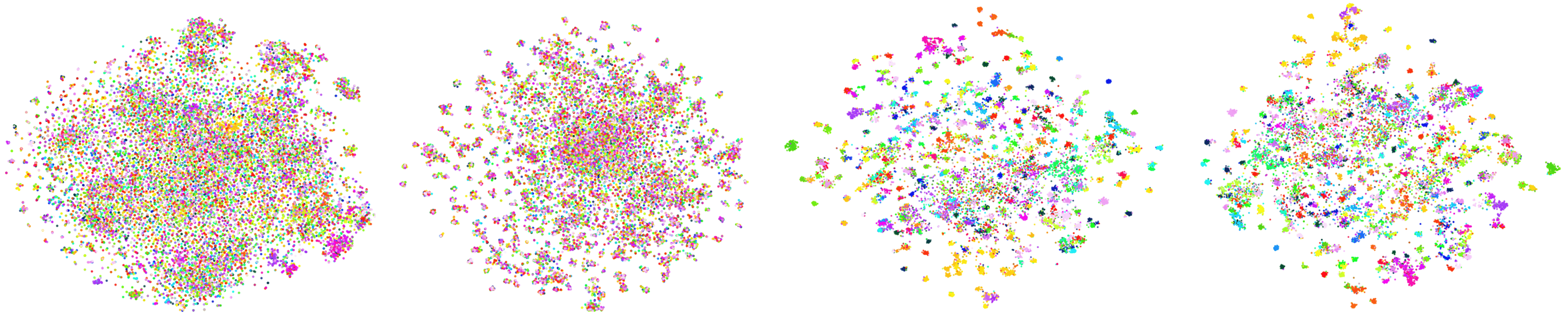}
        \vspace{0.1cm}
        \caption*{(a)}
    \end{minipage}
    \begin{minipage}{0.2\textwidth}
 \includegraphics[width=\linewidth]{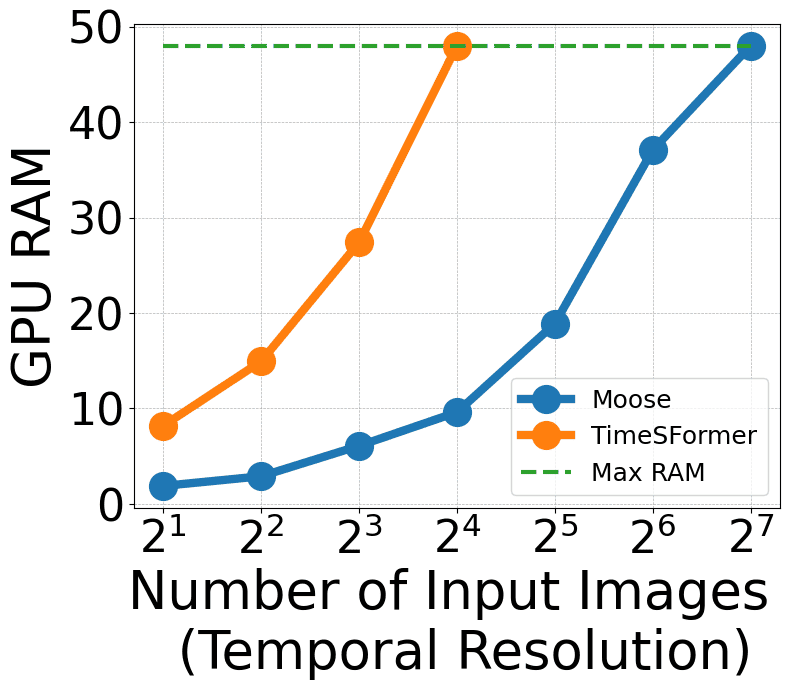}
        \caption*{(b)}
    \end{minipage}
    \begin{minipage}{0.2\textwidth}
        \includegraphics[width=\linewidth]{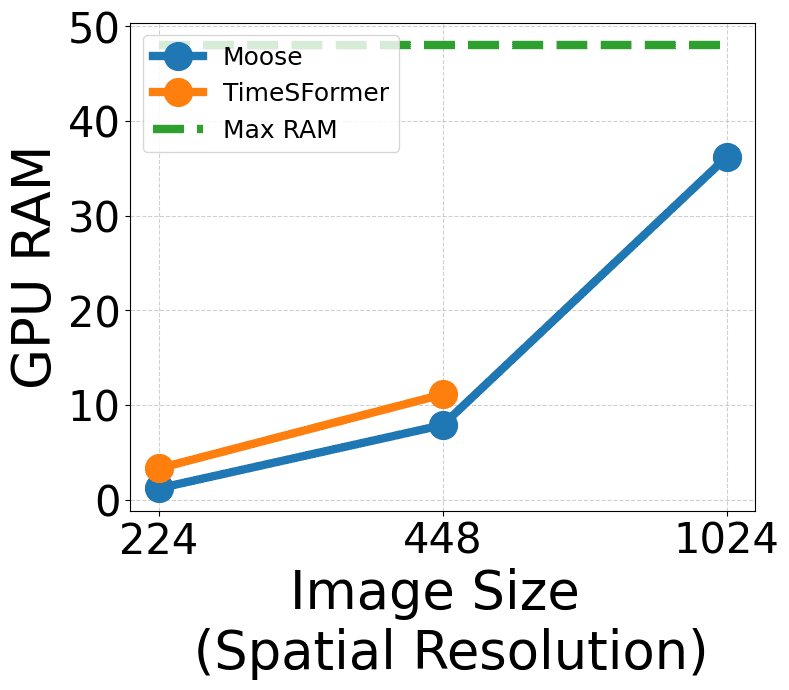}
    \caption*{(c)}
    \end{minipage}
    \caption{Images show (a) TSNE plots of VideoMAE, ViViT, TimeSFormer, MOOSE, respectively and the GPU memory footprint of MOOSE and TimeSFormer during training under varying spatial (b) and temporal (c) resolutions}
\label{fig:tsneplot}
\end{figure}

\noindent
\textbf{Feature Representation via t-SNE.}
To further investigate the quality of learned video representations, we visualize the feature embeddings from different models using t-SNE on the Kinetics-400 (K400) test set Figure \ref{fig:tsneplot}. Each point represents a video sample, and colors denote different action categories. As shown in Figure \ref{fig:tsneplot}(a), the feature clusters from VideoMAE and ViViT appear densely entangled, indicating limited inter-class separability and weaker semantic structure in the learned embeddings. In contrast, TimeSFormer and MOOSE produce significantly more discriminative and well-separated clusters, demonstrating improved class-wise organization.
These visualizations reinforce the quantitative results reported earlier, and confirm MOOSE’s ability to learn more robust and semantically meaningful representations for video understanding.

\subsection{Ablation study}
\textbf{Performance across different fusion modes.}
Table \ref{tab:fusionmodes} shows that the Bidirectional Concat model outperforms all other approaches, achieving 70.84\% top-1 and 89.56\% top-5 accuracy, albeit at a higher cost of 297.3G TFLOPs. This highlights the effectiveness of jointly leveraging both visual and motion cues in a unified representation. In contrast, both Visual-Prior and Motion-Prior Cross-Attention achieve significantly lower top-1 accuracy (55.73–55.75\%), suggesting that relying on a single prior leads to underfitting or suboptimal fusion. Despite similar computational demands, these unimodal-prior approaches fail to match the performance of Bidirectional Concat, highlighting the importance of symmetric, bidirectional integration of spatial and temporal features for robust video understanding.

Figure \ref{fig:vidmodesCAM} shows that bidirectional fusion offers more robust and context-aware spatiotemporal grounding for fine-grained action understanding.  Specifically, the visual-prior attention model predominantly focuses on high-contrast regions of the guitar and the musician's hands, yet exhibits relatively coarse and spatially static activation across frames. The optical flow-prior model, by contrast, demonstrates more temporally aligned and motion-sensitive attention, emphasizing dynamic regions such as strumming fingers and the shifting hand on the fretboard. This configuration captures fine-grained temporal cues, although it may be more susceptible to motion blur or background noise. The bidirectional concat approach integrates both visual and motion features, yielding broader and more stable attention maps that adapt fluidly to both static spatial and temporal movement. 

\textbf{Performance across different temporal aggregation}
Table \ref{tab:tenoaggr} compares the impact of various temporal aggregation methods on video-level accuracy for the Kinetics-400 (K400) dataset. The Causal aggregation approach achieves the highest top-1 and top-5 accuracy, respectively. In contrast, both Mean pooling and Mamba aggregation methods yield substantially lower top-1 accuracy (64.19\%) and comparable top-5 accuracy (88.19\%), despite having identical parameter budgets (101.01G). These results suggest that simple averaging approaches fail to capture the complex temporal dependencies required for fine-grained video understanding. The strong performance of causal aggregation demonstrates the importance of temporal ordering and directional modeling in aggregating frame-level features.

\textbf{Results on other datasets} Table \ref{tab:otherdatasets} presents model performance across four benchmark datasets—rtMRI-Phoneme, AV-ASD, HAA500, and UCF-101—under consistent test conditions (1×1 crop, identical parameter and compute budgets). MOOSE outperforms TimeSFormer on two out of four tasks. Notably, the only dataset where MOOSE underperforms is rtMRI-Phoneme, where it falls short by $7.65$ points. This may be because of the weak optical flow estimation in the challenging (low spatio-temporal resolution) dynamic medical image videos that represent complex organ movements.

\section{Conclusion}
This work addresses the significant challenge of capturing temporal dynamics in large vision-language models for video understanding, a bottleneck where existing attention-based methods struggle with scalability. We introduced MOOSE (Motion Flow Over Spatial Space), a novel video encoder specifically designed for temporally focused tasks. By representing temporal information using optical flow, MOOSE effectively overcomes scalability limitations related to image resolution, sampling rates, and sequence length, while drastically reducing computational demands. Our approach successfully integrates temporal flow into image embeddings, emulating human motion perception. The results demonstrate that MOOSE achieves state-of-the-art performance and maintains robust interpretability, requiring only a minimal number of retrainable parameters. Therefore, MOOSE offers a highly efficient and effective solution for advancing video understanding, particularly in applications demanding strong temporal reasoning.

\noindent
\textbf{Limitations} Although MOOSE effectively leverages a dual-branch attention architecture and incorporates optical-flow information, it still faces several challenges when processing complex videos. First, its temporal resolution for optical-flow computation is limited, so if frames are too sparse or objects move too quickly, the system can suffer from “motion blur” and fail to capture accurate motion vectors. Second, in handheld or drone footage with strong camera shake, background motion noise can cause the attention module to confuse foreground and background, and while applying an image stabilization pre-processing step can mitigate some of this noise, it incurs additional computational overhead. Finally, MOOSE’s generalization capability remains limited when encountering atypical motion patterns or novel camera angles not seen during training, leading to bias and misidentification or omission of important regions. These limitations indicate that future development of MOOSE should focus on enhancing robustness to motion and viewpoint variations to improve accuracy and broaden its applicability.

\noindent
\textbf{Future work} A promising direction for future work is scaling MOOSE for integration into multimodal large language models (MLLMs). Most current MLLMs that handle video do not employ specialized video encoders; instead, they aggregate frame-based image embeddings (often ViT or CLIP models) \cite{videoprism24,videollama,videollama2,videollama3, xu2024slowfast, lin2023video,zhu2023languagebind}.
This frame-wise encoding approach has been popular due to practicality: large image Transformers like CLIP ViT are readily available and have strong representations, so reusing them leverages powerful pretrained knowledge. While practical and efficient, this approach often leads to hallucinated or inaccurate responses for questions involving atomic actions or complex human–object interactions, due to the absence of temporal modeling.
Training a full video transformer from scratch to feed into an LLM would be far more data- and compute-hungry. This brings up a key question: Can we impose an additional temporal pathway to support motion understanding as well as preserve current available spatial encoder?
Another compelling extension is long-form video understanding. Prior approaches have struggled to maintain temporal coherence over extended sequences, relying solely on image embeddings that lack true temporal grounding. In contrast, MOOSE decomposes video into spatiotemporal units, enabling the application of sequence models—such as state-space models or Mamba \cite{mamba, mamba2}—for efficient and scalable temporal aggregation.

\noindent
\textbf{Broader Impact} MOOSE has the potential to greatly benefit society, particularly in clinical and medical domains, where they can enable early detection and analysis of atomic behaviors—such as atypical social interaction or movement patterns—in neurological and developmental conditions (e.g., Autism, Parkinson's disease). However, their deployment raises concerns around privacy, surveillance, and misuse, particularly if applied without informed consent or proper safeguards. Ensuring ethical use and transparency is critical to prevent potential harm, especially in sensitive environments involving vulnerable individuals.

\bibliography{moose}





\newpage
\appendix

\section{Supplemental material}

This section details the experimental setup for training, evaluating, and visualizing the MOOSE model, ensuring transparency and reproducibility. The subsections below describe the datasets, backbones, evaluation metrics, implementation details, baseline models for comparison, and visualization methods. Code is anonymously available at \url{https://anonymous.4open.science/r/MOOSE-C055}

\subsection{More heatmaps}
\begin{figure*}[!h]
\begin{center}
\includegraphics[width=\linewidth]{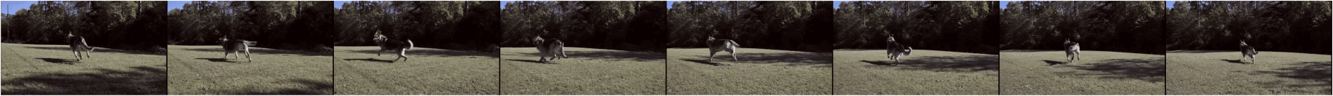} \\
\includegraphics[width=\linewidth]{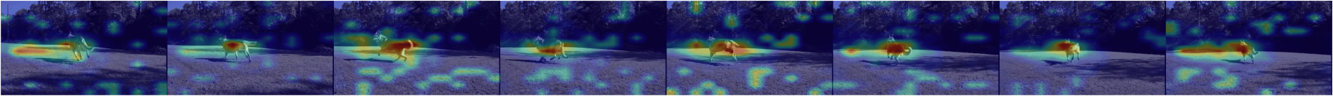} \\
\includegraphics[width=\linewidth]{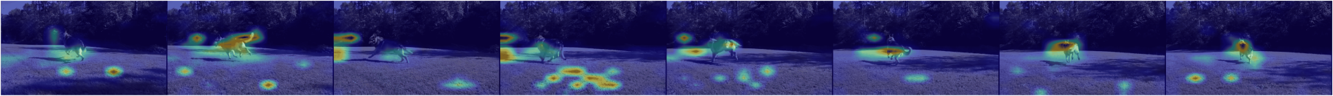} \\
\includegraphics[width=\linewidth]{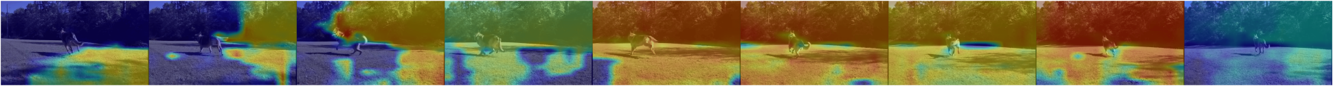}
\end{center}
   \caption{Eight consecutive heatmaps from top to bottom for TimeSformer, ViViT, and MOOSE, respectively, on a moving camera video.}
\label{fig:movecam}
\end{figure*}

\begin{figure*}[!h]
\begin{center}
\includegraphics[width=\linewidth]{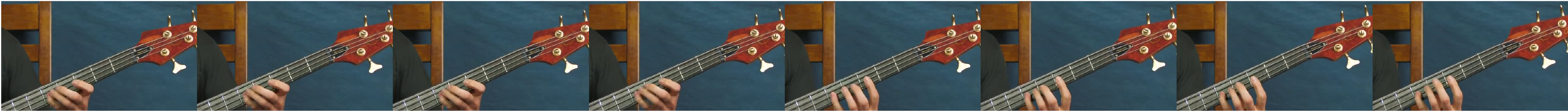} \\
\includegraphics[width=\linewidth]{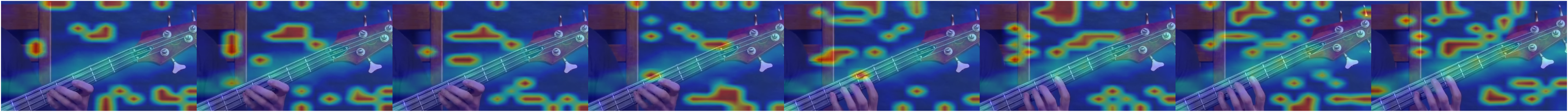} \\
\includegraphics[width=\linewidth]{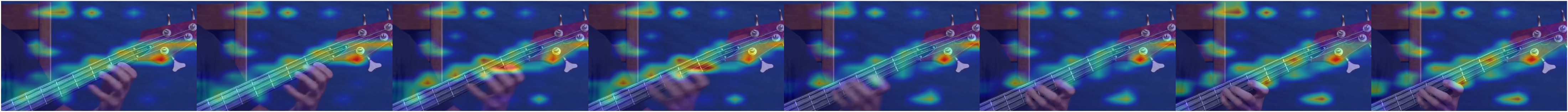} \\
\includegraphics[width=\linewidth]{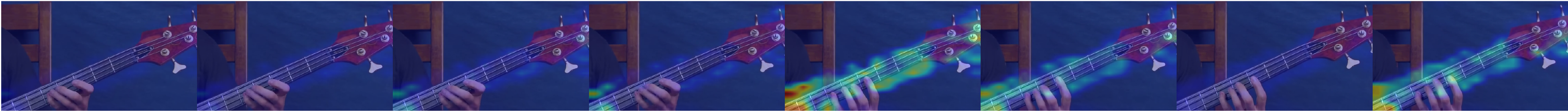}
\end{center}
   \caption{Eight consecutive heatmaps from top to bottom for TimeSformer, ViViT, and MOOSE, respectively, on a static camera video.}
\label{fig:staticcam}
\end{figure*}

Figure \ref{fig:movecam} and \ref{fig:staticcam} visualize eight consecutive heatmaps for three video transformer models — TimeSformer, ViViT, and MOOSE (from top to bottom in each panel) — on two types of videos: a moving camera setup and a static camera setup.
Each row represents the temporal evolution of model attention over eight frames. In the static camera setting, all models exhibit temporally coherent attention, but MOOSE shows more consistent and semantically aligned focus on the object of interest (e.g., the musician’s hand and guitar neck), while TimeSformer and ViViT occasionally attend to less relevant background regions.
In the moving camera case, the performance of MOOSE degrades visibly — attention becomes scattered and more sensitive to global motion, especially across background areas. TimeSFormer and ViViT, by contrast, maintain stable and localized attention, suggesting less dependence on camera moving along with the subject of interest

\begin{figure*}[!h]
\begin{center}
\includegraphics[width=\linewidth]{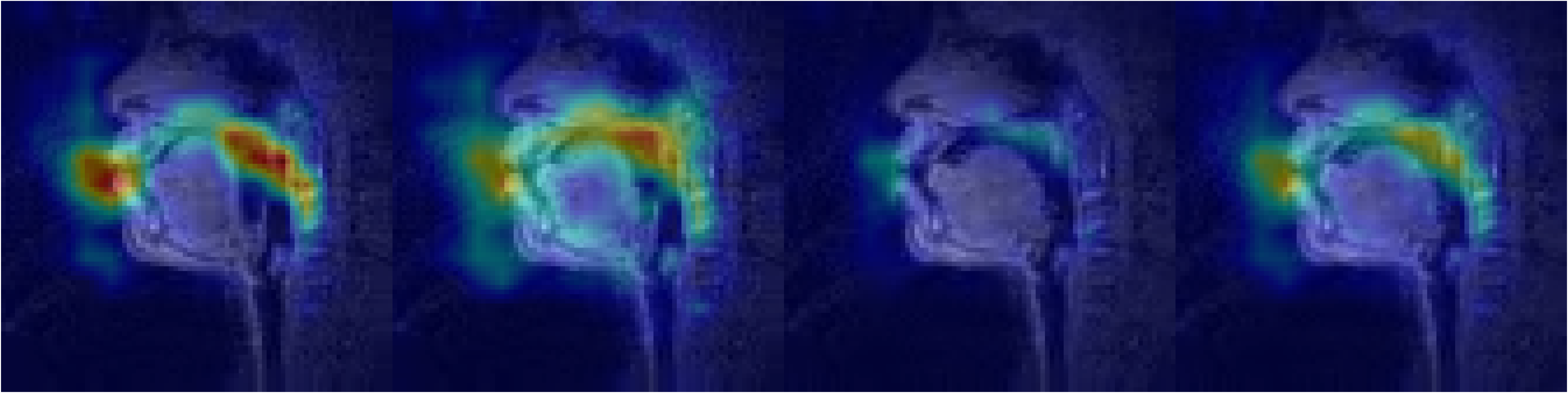} \\
\includegraphics[width=\linewidth]{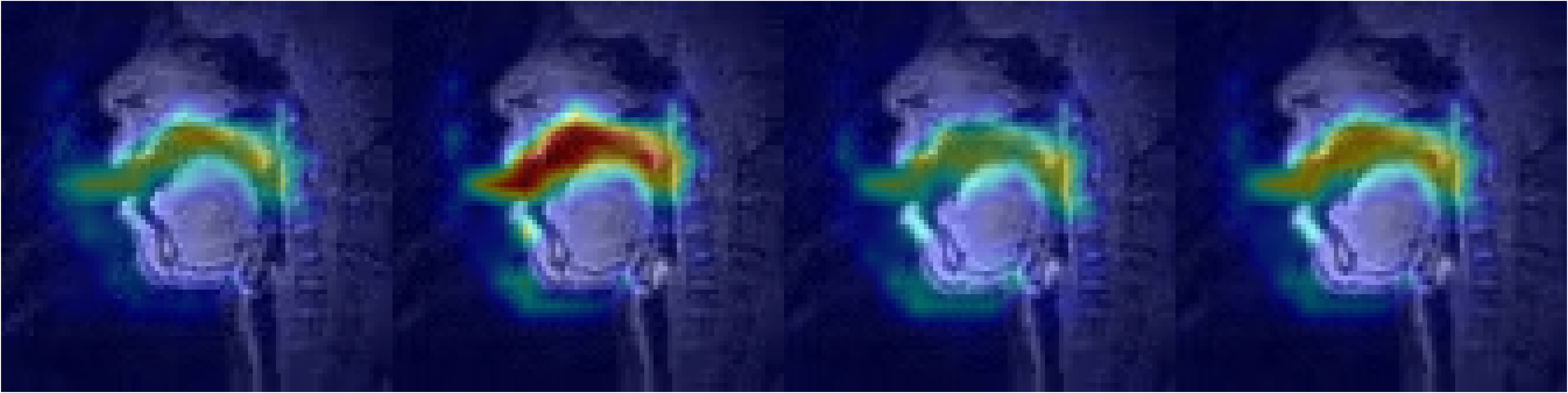} 
\end{center}
   \caption{Attention heatmap of MOOSE on rtMRI-Phoneme dataset. The articulator is producing AA and UW phonemes in the upper and lower four images.}
\label{fig:rtmri}
\end{figure*}

Figure \ref{fig:rtmri} shows a heatmap visualization of the proposed model’s attention across four consecutive frames from a real-time MRI (rtMRI) sequence of the vocal tract during the articulation of the phoneme AA (as in “father”).
The overlaid heatmaps indicate regions of high model activation, highlighting the tongue root, soft palate, and pharyngeal wall as key areas involved in producing this open back vowel. The progression across frames demonstrates the model’s ability to track dynamic articulatory motion, with strong, consistent focus on relevant structures throughout the utterance. Notably, the model maintains attention even during less visually distinct phases, suggesting robustness to MRI noise and anatomical variability.
These results support the model's capacity to identify linguistically meaningful articulator regions in real-time, contributing to interpretable and physiologically informed speech modeling.

\subsection{Datasets}

We train and evaluate MOOSE on six diverse video datasets, including two primary action recognition datasets (\text{Kinetics-400} and \text{Something-Something v2}) and four additional datasets to assess generalization (\text{rtMRI-Phoneme}, \text{AV-ASD}, \text{HAA500}, and \text{UCF-101}). The datasets are described as follows:

\noindent
\textbf{Kinetics-400 (K400)}: Contains approximately 240K+ training videos and 38K+ test videos, classified into 400 coarse-grained actions such as “running” or “jumping.” The actions depend on scene context and objects present.

\noindent
 \textbf{Something-Something v2 (SSv2)}: Includes roughly 220K training videos and 24K test videos, classified into 174 fine-grained actions, such as “pushing something from left to right” or “opening something”, which requires strong temporal reasoning

 \noindent
 \textbf{rtMRI-Phoneme}: 75speaker-rtMRI is a unique corpus comprising 2D sagittal-view real-time MRI (RT-MRI) videos with synchronized audio recordings from 75 subjects engaged in linguistically motivated speech tasks. We manually extract phonemes, including AA, B, IY, K, N, R, T, UW clips from original videos. The rt-MRI videos are original at 86FPS and phonemes have average length of 200ms.

 \noindent
 \textbf{AV-ASD}: The Audio-Visual Autism Spectrum Dataset (AV-ASD) is a curated, multimodal dataset developed to advance research on autism-related behaviors, particularly within social interaction contexts. It comprises 928 annotated clips extracted from 569 publicly available YouTube and Facebook videos, encompassing a wide range of behaviors and real-world environments.

 \noindent
 \textbf{HAA500}: HAA500 is a manually annotated, human-centric dataset for atomic action recognition, encompassing 500 fine-grained action classes with over 591,000 labeled frames. In contrast to prior atomic action datasets that employ coarse-grained, verb-based labels (e.g., “Throw”), HAA500 distinguishes only visually and semantically consistent actions under the same label (e.g., “Baseball Pitching” vs. “Free Throw in Basketball”), thereby reducing ambiguities in action classification. The dataset is meticulously curated to emphasize human figure motion while minimizing spatio-temporal labeling noise, thus improving the training effectiveness of deep neural networks.

 \noindent
 \textbf{UCF-101}: UCF101 is a large-scale action recognition dataset comprising 13,320 videos spanning 101 distinct action categories, sourced from YouTube to reflect real-world scenarios. Videos are organized into 25 groups per action category, with each group containing 4–7 clips that may share contextual similarities such as background or viewpoint. The action categories are broadly categorized into five types: Human-Object Interaction, Body-Motion Only, Human-Human Interaction, Playing Musical Instruments, and Sports.

Each video is sampled at rate 1/8 meaning that for every 8 frame, sample one image. For our model settings, the input size for most datasets is also 9 (8+1) frames, the plus one frames is used for optical flow extraction. Because datasets may have different FPS and durations, rtMRI is sample at rate 1/4 and number of input frames is 5 (4+1). Frames are resized to $224 \times 224$ pixels for preprocessing (unless otherwise specified, e.g., $448 \times 448$ for high-resolution tests). Detailed information on the primary datasets is given in \text{Table~\ref{tab:datasets}}.

\begin{table}[h]
    \centering
    \caption{Overview of the primary datasets.}
    \label{tab:datasets}
    \begin{tabular}{lcccccc}
        \toprule
        Dataset & \# Train & \# Test & \# Val  & \# Classes & Video FPS & Average duration\\
        \midrule
        K400 & 240K & 38K & 20K & 400 & 12, 24, 30 &10s\\
        SSv2 & 220K & 24K & 24K & 174 & 12 & 4s\\
        rtMRI-Phoneme & 40K & 20K & 15K & 8 & 86 & 200ms\\
        HAA500 & 4K & 3K & 38K & 500 & 30 &1s\\
        AV-ASD & 600 &150 &150 &11 &25 &30s\\
        UCF-101 & 7K & 3K &3K & 101 &30 &3s \\
        \bottomrule
    \end{tabular}
\end{table}

\subsection{Backbones: Vision Transformer (ViT) and RAFT}

The MOOSE model uses three main backbones:

\noindent
\textbf{Vision Transformer for spatial embeddings}: A pretrained ViT with DINOv2 (\texttt{ViT-B/14 distilled} with register), having 12 layers and an embedding dimension of 768. T Input video frames are split into $14 \times 14$ patches to form spatial embeddings with 3 input channel, number of head 12.

\noindent
\textbf{RAFT}: An optical flow estimation model pretrained on the Sintel dataset. RAFT takes consecutive frame pairs and outputs a flow field, which is treated as an input “image” for the temporal ViT.

\noindent
\textbf{Vision Transformer for optical flow embeddings} We use a smaller ViTs for flows embeddings. Consider extracted optical flow as 2-channel image, we use default tiny ViTs from DINOv2 having 12 layers with 2 input channel, patch size 14, embeddings dimension 192, number of head 3. Notice that the embeddings dimension of ViTs for spatial much smaller than temporal, this was chosen to reduce computational on optical flows. The ViT for temporal resolution was trained from scratch

Only the temporal ViT, the classification head, and the fusion module are trained; the spatial ViT and RAFT remain frozen.

\subsection{Metrics}

MOOSE’s performance is evaluated using the following metrics:

\noindent
\textbf{Top-1 Accuracy}: The fraction of samples for which the highest-scoring prediction matches the ground truth:
    \begin{equation}
        \text{Top-1 Accuracy} = \frac{|\{i \mid \hat{y}_i = y_i\}|}{N}
    \end{equation}
    where $\hat{y}_i$ is the top prediction for sample $i$, $y_i$ is the true label, and $N$ is the total number of samples.

\noindent
\textbf{Top-5 Accuracy}: The fraction of samples for which the true label is among the top five predictions:
    \begin{equation}
        \text{Top-5 Accuracy} = \frac{|\{i \mid y_i \in \{\hat{y}_{i,1}, \ldots, \hat{y}_{i,5}\}\}|}{N}
    \end{equation}
    where $\{\hat{y}_{i,1}, \ldots, \hat{y}_{i,5}\}$ are the five highest-probability predictions.

\noindent
\textbf{TFLOPs}: The number of floating-point operations per second, measuring computational complexity. For MOOSE, TFLOPs is computed over all layers in ViT and RAFT:
    \begin{equation}
        \text{TFLOPs} = \sum_{\text{layers}} (\text{FLOPs}_{\text{attention}} + \text{FLOPs}_{\text{feedforward}} + \text{FLOPs}_{\text{RAFT}})
    \end{equation}
    MOOSE achieves 297.3G TFLOPs for the two-dimensional fusion approach.

\noindent
\textbf{GPU Hours}: Training time normalized to NVIDIA Tesla V100 GPUs:
    \begin{equation}
        \text{GPU Hours} = T_{\text{train}} \times N_{\text{GPU}} \times \alpha_\text{norm}
    \end{equation}
where $T_{\text{train}}$ is the actual training duration and $N_{\text{GPU}}$ is the number of GPUs (1 in this case) and $\alpha_\text{norm}$ normalized approximate relative compute ratios: V100 $\approx$ 1$\times$, A6000 $\approx$ 1.4$\times$, A100 $\approx$ 2.5$\times$. MOOSE requires about 132 GPU hours on K400.

All metrics are averaged over three runs with different random seeds.

\subsection{Implementation Details}

Experiments with K400, SSv2, AV-ASD are run on an NVIDIA A6000 GPU (48\,GB RAM) and other dataset are trained on GTX 3090 (24\,GB RAM) using Python 3.10 in the \texttt{Conda} environment. Key implementation details include:

\noindent
\textbf{Batch Size}: 16 for experiments with GTX 3090 and 96 for experiments with A6000. All Moose's experiments trained in 20 epochs.

\noindent
\textbf{Optimizer}: Stochastic Gradient Descent (SGD) with momentum 0.9 and weight decay of $1\times10^{-4}$.

\noindent
\textbf{Learning Rate}: Initialized learning rate at 0.005, adjusted by cosine annealing:
    \begin{equation}
        \eta_t = \eta_{\min} + \frac{1}{2} (\eta_{\max} - \eta_{\min}) \Bigl(1 + \cos\bigl(\tfrac{t}{T_{\max}}\pi\bigr)\Bigr)
    \end{equation}
    where $\eta_t$ is the learning rate at step $t$, $\eta_{\max}=0.001$, $\eta_{\min}=0$, and $T_{\max}$ is the total number of epochs.
    
\noindent
\textbf{Data Augmentation}: Random cropping and horizontal flipping.

\noindent
\textbf{Early Stopping}: Training halts if no improvement is seen for 10 epochs.

\noindent
\textbf{Input Configuration}: 8 frames sampled at 8× rate, resolution $224\times224$ (or $448\times448$ for high-resolution tests).

\subsection{Visualize TimeSFormer, ViViT and Moose Attention Heatmap}
MOOSE’s 132 GPU hours on K400 is substantially lower than SlowFast’s 1188 GPU hours, and it uses over 30\% less GPU memory than TimeSFormer at $448\times448$ resolution. We show below algothrms to plot attention heatmap of ViViT, TimeSFormer, and Moose

\begin{algorithm}[ht]
\caption{TimeSformer's Attention Heatmap Visualization}
\begin{algorithmic}[1]
\Procedure{vizTimeSFormerAttention}{}
    \State $\texttt{frames} \gets \text{loadVideo}\left(\text{\textcolor{Blue}{``path/to/video''}}\right)$
    \State $\texttt{model} \gets \text{loadTimeSformer}\left(\text{\textcolor{Blue}{``path/to/checkpoint''}}\right)$
    \State $\texttt{inputs} \gets \text{processFrames}(\texttt{frames})$
    \State $\texttt{heatMaps} \gets [\text{None}]$
    \For{$i \gets 0 \text{ to } \text{Batch}-1$}
        \State $\texttt{attn} \gets \texttt{model}(\texttt{inputs[i]}).\text{ViTEncoder}.\text{layer}\texttt{[-1]}.\text{attentions}\texttt{[-1]}$ \\ \Comment{\texttt{attn}.shape $\equiv$ [Head, Patches + Cls, Patches + Cls]}
        \State $\texttt{attMap} \gets \text{reshape}\left(\text{norm}\left(\text{mean}\left(\texttt{attn}\left([i, :, 0, 1:\right], \text{dim} = 0\right)\right), (14, 14)\right)$
        \State $\texttt{Heatmaps[i]} \gets \text{interpolate}\left(\texttt{attMap}, (W, H)\right)$
        \Comment{\texttt{Heatmaps}.shape $\equiv$ [T, W, H]}
    \EndFor
    \State $\texttt{Outputs} \gets \text{overlayHeatmaps}(\texttt{frames}, \texttt{heatMaps}, \alpha)$
    \State \Return $\texttt{Outputs}$
\EndProcedure
\end{algorithmic}
\end{algorithm}

\newpage
\begin{algorithm}[ht]
\caption{ViViT's Attention Heatmap Visualization}
\begin{algorithmic}[1]
\Procedure{vizViViTAttention}{}
    \State $\texttt{frames} \gets \text{loadVideo}\left(\text{\textcolor{Blue}{``path/to/video''}}\right)$
    \State $\texttt{model} \gets \text{loadViViT}\left(\text{\textcolor{Blue}{``google/vivit-b-16x2-kinetics400''}}\right)$
    \State $\texttt{inputs} \gets \text{processFrames}(\texttt{frames})$
    \State $\texttt{heatMaps} \gets [\text{None}]$
    \State $\texttt{attn} \gets \texttt{model}(\texttt{inputs}).\text{ViTEncoder}.\text{layer}[-1].\text{attentions}[-1]$ 
    \For{$t \gets 0 \text{ to } T-1$} \Comment{T: number of tubelets}
        \State $\texttt{attMap} \gets \text{reshape}\left(\text{norm}\left(\text{mean}\left(\texttt{attn}[0, :, 0, 1 + t \cdot P : 1 + (t+1) \cdot P], \text{dim}=0\right)\right)\right)$ \\ \Comment{\texttt{attMap}.shape $\equiv$ [14, 14]}
        \State $\texttt{heatMaps}[t] \gets \text{interpolate}\left(\texttt{attMap}, (W, H)\right)$ \Comment{\texttt{heatMaps}.shape $\equiv$ [T, W, H]}
    \EndFor
    \State $\texttt{Outputs} \gets \text{overlayHeatmaps}(\texttt{frames}, \texttt{heatMaps}, \alpha)$
    \State \Return $\texttt{Outputs}$
\EndProcedure
\end{algorithmic}
\end{algorithm}

\begin{algorithm}[ht]
\caption{MOOSE's Attention Heatmap and Optical Flow Visualization}
\begin{algorithmic}[1]
\Procedure{vizAttentionHeatmap}{}
    \State $\texttt{frames} \gets \text{loadVideo}\left(\text{\textcolor{Blue}{``path/to/video''}}\right)$
    \State $\texttt{model} \gets \text{loadMOOSE}\left(\text{\textcolor{Blue}{``path/to/checkpoint''}}\right)$
    \State $\texttt{inputs} \gets \text{processFrames}(\texttt{frames})$
    \State $\texttt{q}_\text{flow}, \texttt{k}_\text{flow} \gets \texttt{model}(\texttt{inputs}). \text{joint\_cross\_attn.to\_qk}$
    \State $\texttt{q}, \texttt{k} \gets \texttt{model}(\texttt{inputs}). \text{joint\_cross\_attn.context\_to\_qk}$
    \State $\texttt{heatMaps} \gets [\text{None}]$
    \State $\texttt{attn\_spatial}, \texttt{attn\_temporal} \gets \text{softmax}\left(\frac{\texttt{q} \cdot \texttt{k}^\top}{\sqrt{d}}\right), \text{softmax}\left(\frac{\texttt{q}_\text{flow} \cdot \texttt{k}_\text{flow}^\top}{\sqrt{d}}\right)$ 
    \For{$i \gets 0 \text{ to } B-1$}
        \State $\texttt{attMap} \gets \text{reshape}\left(\text{norm}\left(\texttt{attn\_spatial}[i, 1:, 0]\right), (16, 16)\right)$ \\ \Comment{\texttt{attMap}.shape $\equiv$ [16, 16]}
        \State $\texttt{heatMaps}[i] \gets \text{interpolate}\left(\texttt{attMap}, (W, H)\right)$ \Comment{\texttt{heatMaps}.shape $\equiv$ [B, W, H]}
    \EndFor
    \State $\texttt{flowMaps} \gets [\text{None}]$
    \For{$i \gets 0 \text{ to } B-2$}
        \State $\texttt{flow} \gets \text{RAFT}(\texttt{frames}[i], \texttt{frames}[i+1])$ \Comment{\texttt{flow}.shape $\equiv$ [2, H, W]}
        \State $\texttt{arrows} \gets \text{sampleArrows}(\texttt{flow}, \text{grid\_step}, \text{scale})$ \Comment{Sampled flow vectors} 
        \State $\texttt{attMap} \gets \text{reshape}\left(\text{norm}\left(\texttt{attn\_temporal}[i, 1:, 0]\right), (16, 16)\right)$
        \State $\texttt{flowheatMaps}[i] \gets \text{interpolate}\left(\texttt{attMap}, (W, H)\right)$\\ \Comment{\texttt{flowMaps}.shape $\equiv$ [B-1, H, W, 3]}
        \State $\texttt{flowMaps}[i] \gets \text{overlayArrows}(\texttt{arrows}, \texttt{flowheatMaps}[i])$
    \EndFor
    \State \Return $\texttt{flowMaps}$, $\texttt{heatMaps}$
\EndProcedure
\end{algorithmic}
\end{algorithm}

\end{document}